\documentclass[11pt,a4paper]{article}
\usepackage[hyperref]{acl2019}
\usepackage{times}
\usepackage{latexsym}

\usepackage{url}

\usepackage{graphicx}  %Required
\usepackage{latexsym}
\usepackage{enumitem}
\usepackage{stfloats}
\usepackage{float}
\usepackage{nccmath}
\usepackage{mathtools}
\usepackage{epstopdf}
\usepackage{amsfonts}
\usepackage{amssymb}
\usepackage{CJKutf8}
\usepackage{CJK}
\usepackage{amsmath}
\usepackage{verbatim}
\usepackage{multirow}
\usepackage{subfigure}
\usepackage{pifont}% http://ctan.org/pkg/pifont
\usepackage{array}
\usepackage{fp}
\usepackage{makecell}
\usepackage{cases}
\usepackage{makecell}
\usepackage{color, soul}
\usepackage{xcolor}
\usepackage{bm}
\usepackage{booktabs}
\usepackage{tablefootnote}
\usepackage{tcolorbox}
\usepackage[linesnumbered,ruled,vlined]{algorithm2e}
\usepackage[noend]{algpseudocode}

\usepackage{url}
\usepackage{graphicx}  %Required
\usepackage{amsfonts}
\usepackage{amsmath}
\usepackage{bm}
\usepackage{mathrsfs}
\usepackage{float}
\usepackage{colortbl}

\usepackage{subfigure}
\usepackage{booktabs,multirow}
\usepackage{bigstrut,bigdelim}
\usepackage{paralist}
\usepackage{bbm}
\usepackage{diagbox}

\aclfinalcopy % Uncomment this line for the final submission
\def\aclpaperid{2554p} %  Enter the acl Paper ID here

\title{Neural Response Generation with Meta-Words}

\author{Can Xu$^{1}$, Wei Wu$^{1}$\thanks{\ \ Corresponding author.}\ \ , Chongyang Tao$^{2}$, Huang Hu$^{1}$, Matt Schuerman$^{3}$, and Ying Wang$^{3}$\\
$^1$Microsoft Corporation, Beijing, China\\
$^2$Institute of Computer Science and Technology, Peking University, Beijing, China \\
$^3$Microsoft Corporation, Redmond, Washington\\
{\tt $^{1,3}$\{caxu,wuwei,huahu,matthes,Ying.Wang\}@microsoft.com} \\
{\tt $^{2}$chongyangtao@pku.edu.cn } \\}

\date{}

\newtcbox{\mybox}[1][red]
{on line, arc = 0pt, outer arc = 0pt,
colback = #1!10!white, colframe = #1!50!black,
boxsep = 0pt, left = 0pt, right = 0pt, top = 2pt, bottom = 2pt,
boxrule = 0pt, bottomrule = 0.5pt, toprule = 0.5pt}
\begin{document}
\maketitle
\begin{abstract}
We present open domain response generation with meta-words. A meta-word is a structured record that describes various attributes of a response, and thus allows us to explicitly model the one-to-many relationship within open domain dialogues and perform response generation in an explainable and controllable manner.  To incorporate meta-words into generation, we enhance the sequence-to-sequence architecture with a goal tracking memory network that formalizes meta-word expression as a goal and manages the generation process to achieve the goal with a state memory panel and a state controller. Experimental results on two large-scale datasets indicate that our model can significantly outperform several state-of-the-art generation models in terms of response relevance, response diversity, accuracy of one-to-many modeling, accuracy of meta-word expression, and human evaluation.   
\end{abstract}
\section{Introduction}
Human-machine conversation is a fundamental problem in NLP. Traditional research focuses on building task-oriented dialog systems \cite{young2013pomdp} to achieve specific user goals such as restaurant reservation through limited turns of dialogues within specific domains. Recently, building a chatbot for open domain conversation \cite{vinyals2015neural} has attracted increasing attention, not only owing to the availability of large amount of human-human conversation data on internet, but also because of the success of such systems in real products  such as the social bot XiaoIce \cite{shum2018eliza} from Microsoft.  

A common approach to implementing a chatbot is to learn a response generation model within an encoder-decoder framework \cite{vinyals2015neural,shangL2015neural}. Although the architecture can naturally model the correspondence between a message and a response, and is easy to extend to handle conversation history \cite{serban2015building,xing2017hierarchical} and various constraints \cite{li2016persona,zhou2017emotional}, it is notorious for generating safe responses such as ``I don't know'' and ``me too'' in practice. A plausible reason for the ``safe response'' issue is that there exists one-to-many relationship between messages and responses. One message could correspond to many valid responses and vice versa \cite{zhang2018learning}. The vanilla encoder-decoder architecture is prone to memorize high-frequency patterns in data, and thus tends to generate similar and trivial responses for different messages. A typical method for modeling the relationship between messages and responses is to introduce latent variables into the encoder-decoder framework \cite{serban2017hierarchical,zhao2017learning,park2018hierarchical}. It is, however,  difficult to explain what relationship a latent variable represents, nor one can control responses to generate by manipulating the latent variable. Although a recent study \cite{zhao2018unsupervised} replaces continuous latent variables with discrete ones, it still needs a lot of post human effort to explain the meaning of the variables.   

\begin{table}[!t] 
	\centering
	\resizebox{0.48\textwidth}{!}{
    \begin{tabular}{|rl|}
		\hline
		\textbf{Message:} & last week I have a nice trip to New York! \\ \hline
		\textbf{Meta-word:} & {\color{blue} Act: yes-no question $\vert$ Len: 8 $\vert$ Copy: true $\vert$ Utts: false $\vert$ Spe: medium} \\
		\textbf{Response 1:} & Is \underline{New York} more expensive than California? \\ \hline
		
		\textbf{Meta-word:} &  {\color{blue} Act: wh-question $\vert$ Len: 17 $\vert$ Copy: false $\vert$ Utts: true $\vert$ Spe: high} \\
		\textbf{Response 2:} & Cool, sounds great! What is the tallest building in this city, \textbf{Chrysler} building?\\ \hline

		\textbf{Meta-word:} & {\color{blue} Act: statement $\vert$ Len: 13 $\vert$ Copy: false $\vert$ Utts: true $\vert$ Spe: low} \\  
		\textbf{Response 3:} & I don't know what you are talking about. But it seems good. \\ \hline
	\end{tabular}
	}
	%\vspace{-2mm}
	\caption{An example of response generation with meta-words. The \underline{underlined word} means it is copied from the message, and the \textbf{word in bold} means it corresponds to high specificity. } 
	%\vspace{-7mm}
\label{example}
\end{table}

In this work, we aim to model the one-to-many relationship in open domain dialogues in an explainable and controllable way.  Instead of using latent variables, we consider \textit{explicitly} representing the relationship between a message and a response with meta-words\footnote{We start from single messages. It is easy to extend the proposed approach to handle conversation history. }. A meta-word is a structured record that characterizes the response to generate. The record consists of a group of variables with each  an attribute of the response. Each variable is in a form of (key, type, value) where ``key'' defines the attribute, ``value'' specifies the attribute, and ``type'' $\in \{r,c\}$ indicates whether the variable is real-valued ($r$) or categorical ($c$). Given a message, a meta-word corresponds to one kind of relationship between the message and a response, and by manipulating the meta-word (e.g., values of variables or combination of variables), one can control responses in a broad way. Table \ref{example} gives an example of response generation with various meta-words, where ``Act'', ``Len'', ``Copy'', ``Utts'', and ``Spe'' are variables of a meta-word and refer to dialogue act, response length (including punctuation marks), if copy from the message, if made up of multiple utterances, and  specificity level \cite{zhang2018learning} respectively\footnote{For ease of understanding, we transformed ``copy ratio'' and ``specificity'' used in our experiments into categorical variables.}.  Advantages of response generation with meta-words are three-folds: (1) the generation model is explainable as the meta-words inform the model, developers, and even end users what responses they will have before the responses are generated; (2) the generation process is controllable. The meta-word system acts as an interface that allows developers to customize responses by tailoring the set of attributes; (3) the generation approach is general. By taking dialogue acts \cite{zhao2017learning}, personas \cite{li2016persona}, emotions \cite{zhou2017emotional}, and specificity \cite{zhang2018learning} as attributes, our approach can address the problems in the existing literature in a unified form; and (4) generation-based open domain dialogue systems now become scalable, since the model supports feature engineering on meta-words. 

The challenge of response generation with meta-words lies in how to simultaneously ensure relevance of a response to the message and fidelity of the response to the meta-word. To tackle the challenge, we propose equipping the vanilla sequence-to-sequence architecture with a novel goal tracking memory network (GTMN) and crafting a new loss item for learning GTMN. GTMN sets meta-word expression as a goal of generation and dynamically monitors expression of each variable in the meta-word during the decoding process. Specifically, GTMN consists of a state memory panel and a state controller where the former records status of meta-word expression and the latter manages information exchange between the state memory panel and the decoder.  
In decoding, the state controller updates the state memory panel according to the generated sequence, and reads out difference vectors that represent the residual of the meta-word. The next word from the decoder is predicted based on attention on the message representations, attention on the difference vectors, and the word predicted in the last step. In learning, besides the negative log likelihood, we further propose minimizing a state update loss that can directly supervise the learning of the memory network under the ground truth. We also propose a meta-word prediction method to make the proposed approach complete in practice.

We test the proposed model on two large-scale open domain conversation datasets built from Twitter and Reddit, and compare the model with several state-of-the-art generation models in terms of response relevance, response diversity, accuracy of one-to-many modeling, accuracy of meta-word expression, and human judgment. Evaluation results indicate that our model can significantly outperform the baseline models over most of the metrics on both datasets. 

Our contributions in this paper are three-folds: (1) proposal of explicitly modeling one-to-many relationship and explicitly controlling response generation in open domain dialogues with multiple variables (a.k.a., meta-word); (2) proposal of a goal tracking memory network that naturally allows a meta-word to guide response generation; and (3) empirical verification of the effectiveness of the proposed model on two large-scale datasets.

%\vspace{-1.5mm}
\section{Related Work}
%\vspace{-1.5mm}
Neural response generation models are built upon the encoder-decoder framework \cite{sutskever2014sequence}. Starting from the basic sequence-to-sequence with attention architecture \cite{vinyals2015neural,shangL2015neural}, extensions under the framework have been made to combat the ``safe response'' problem \cite{mou2016sequence,tao2018get}; to model the hierarchy of conversation history \cite{serban2015building,serban2017hierarchical,xing2017hierarchical}; to generate responses with specific personas or emotions \cite{li2016persona,zhou2017emotional}; and to speed up response decoding \cite{wu2017neural}. In this work, we also aim to tackle the ``safe response'' problem, but in an explainable, controllable, and general way. Rather than learning with a different objective (e.g., \cite{li2015diversity}), generation from latent variables (e.g., \cite{zhao2017learning}), or introducing extra content (e.g., \cite{xing2017topic}),  we explicitly describe relationship between message-response pairs by defining meta-words and express the meta-words in responses through a goal tracking memory network. Our method allows developers to manipulate the generation process by playing with the meta-words and provides a general solution to response generation with specific attributes such as dialogue acts.

Recently, controlling specific aspects in text generation is drawing increasing attention \cite{hu2017toward,logeswaran2018content}. In the context of dialogue generation, \citet{wang2017steering} propose steering response style and topic with human provided topic hints and fine-tuning on small scenting data; \citet{zhang2018learning} propose learning to control specificity of responses; and very recently, \citet{see2019makes} investigate how controllable attributes of responses affect human engagement with methods of conditional training and weighted decoding. Our work is different in that (1) rather than  playing with a single variable like specificity or topics, our model simultaneously controls multiple variables and can take controlling with specificity or topics as special cases; and (2) we manage attribute expression in response generation with a principled approach rather than simple heuristics like in \cite{see2019makes}, and thus, our model can achieve better accuracy in terms of attribute expression in generated responses.

%\vspace{-1.5mm}
\section{Problem Formalization}
%\vspace{-1.5mm}
Suppose that we have a dataset $\mathcal{D}=\{(X_i,M_i,Y_i)\}^N_{i=1}$, where $X_i$ is a message, $Y_i$ is a response, and $M_i$ = $(m_{i,1},\ldots, m_{i,l})$ is a meta-word with $m_{i,j}=(m_{i,j}.k, m_{i,j}.t, m_{i,j}.v)$ the $j$-th variable and $m_{i,j}.k$, $m_{i,j}.t$, and $m_{i,j}.v$ the key, the type, and the value of the variable respectively.  Our goal is to estimate a generation probability $P(Y| X,M)$ from $\mathcal{D}$, and thus given a new message $X$ with a pre-defined meta-word $M$, one can generate responses for $X$ according to $P(Y| X,M)$. In this work, we assume that $M$ is given as input for response generation. Later, we will describe how to obtain $M$ with $X$.

\begin{figure*}[t!]
    \centering
    \includegraphics[width=0.88\textwidth]{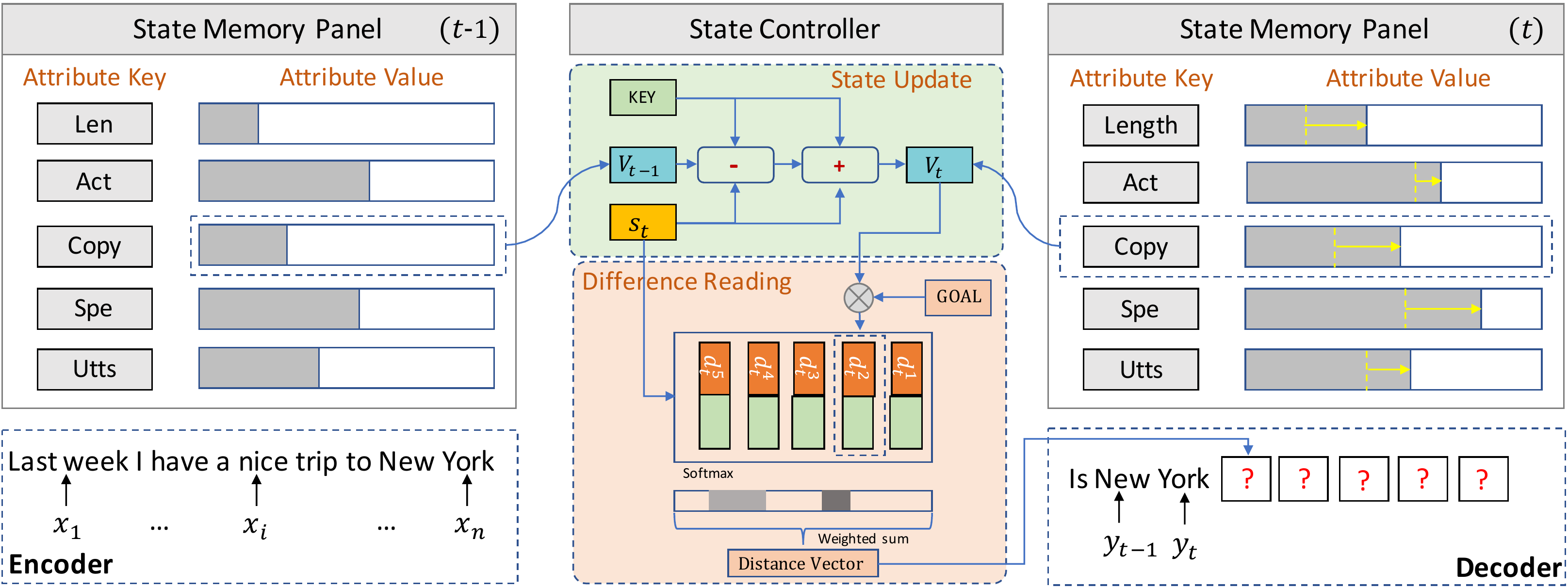}
    \caption{Architecture of goal tracking memory enhanced sequence-to-sequence model.}
    \label{fig:model}
\end{figure*}

%\vspace{-1.5mm}
\section{Response Generation with Meta-Words}
%\vspace{-1.5mm}
In this section, we present our model for response generation with meta-words. We start from an overview of the model, and then dive into details of the goal tracking memory enhanced decoding. 

%\vspace{-1mm}
\subsection{Model Overview}
Figure \ref{fig:model} illustrates the architecture of our goal tracking memory enhanced sequence-to-sequence model (GTMES2S). The model equips the encoder-decoder structure with a goal tracking memory network that comprises a state memory panel and a state controller. 
Before response decoding, the encoder represents an input message as a hidden sequence through a bi-directional recurrent neural network with gated recurrent units (biGRU) \cite{chung2014empirical}, and the goal tracking memory network is initialized by a  meta-word. Then, during response decoding, the state memory panel tracks expression of the meta-word and gets updated by the state controller. The state controller manages the process of decoding at each step by reading out the status of meta-word expression from the state memory panel and informing the decoder of the difference between the status and the target of meta-word expression. Based on the message representation, the information provided by the state controller, and the generated word sequence, the decoder predicts the next word of the response. 

In the following section, we will elaborate the goal tracking memory enhanced decoding, which is the key to having a response that is relevant to the message and at the same time accurately reflects the meta-word.

%\vspace{-1mm}
\subsection{Goal Tracking Memory Network}\label{GTMN}

The goal tracking memory network (GTMN) dynamically controls response generation according to the given meta-word via cooperation of the state memory panel and the state controller. It informs the decoder at the first time to what extend the meta-word has been expressed. For local attributes such as response length\footnote{Local attributes refer to the attributes whose values are location sensitive during response generation. For example, length of the remaining sequence varies after each step of decoding.  In contrary, some attributes, such as dialogue acts, are global attributes, as they are reflected by the entire response.}, the dynamic control strategy is more reasonable than static strategies such as feeding the embedding of attributes to the decoder like in conditional training in \cite{see2019makes}. This is because if the goal is to generate a response with $5$ words and $2$ words have been decoded, then the decoder needs to know that there are $3$ words left rather than always memorizing that $5$ words should be generated. 

%\vspace{-1mm}
\subsubsection{State Memory Panel}
Suppose that the given meta-word $M$ consists of $l$ variables, then the state memory panel $\mathcal{M}$ is made up of $l$ memory cells $\{\mathcal{M}_i\}_{i=1}^l$ where $\forall i\in \{1,\ldots,l\}$, $\mathcal{M}_i$ is in a form of (key, goal, value) which are denoted as 
$\mathcal{M}_i.k$, $\mathcal{M}_i.g$, and $\mathcal{M}_i.v$ respectively. We define $Rep(\cdot)$ as a representation getting function which can be formulated as 
%\vspace{-1mm}
\begin{equation} 
\small
\label{Repfunc}
\begin{aligned}
 \textstyle{Rep(m_i.k)} & = B(m_i.k), \\
Rep(m_i.v) & =\left\{
\begin{aligned}
 &\sigma(B(m_i.v)) ,\quad \quad \quad m_i.t = c\\
 & m_i.v \times \sigma(B(m_i.k)) , m_i.t = r,
\end{aligned}
\right.
\end{aligned}
\end{equation}
where $m_i$ is the $i$-th variable of $M$, $\sigma(\cdot)$ is a sigmoid function, and $B(\cdot)$ returns the bag-of-words representation for a piece of text. $\mathcal{M}_i$ is then initialized as:
%\vspace{-1mm}
\begin{equation}
\small
\label{MIni}
%\left\{
\begin{aligned}
 &\mathcal{M}_i.k = \textstyle{Rep(m_i.k)},\\
 &\mathcal{M}_i.g = \textstyle{Rep(m_i.v)},\\
 &\mathcal{M}_i.v_0 = \mathbf{0}.
 \end{aligned}
%\right.
%\vspace{-1mm}
\end{equation}

$\mathcal{M}_i.k \in \mathbb{R}^d$ stores the key of $m_i$, and $\mathcal{M}_i.g \in \mathbb{R}^d$ stores the goal for expression of $m_i$ in generation. Thus, the two items are frozen in decoding. $\mathcal{M}_i.v \in \mathbb{R}^d$ refers to the gray part of the progress bar in Figure \ref{fig:model}, and represents the progress of expression of $m_i$ in decoding. Hence, it is updated by the state controller after each step of decoding.

%\vspace{-1mm}
\subsubsection{State Controller}
As illustrated by Figure \ref{fig:model}, the state controller stays between the encoder and the decoder, and manages the interaction between the state memory panel and the decoder. Let $s_t$ be the hidden state of the decoder at step $t$. The state controller first updates $\mathcal{M}_i.v_{t-1}$ to $\mathcal{M}_i.v_{t}$ based on $s_t$ with a state update operation. It then obtains the difference between $\mathcal{M}_i.g$ and $\mathcal{M}_i.v_{t}$ from the state memory panel via a difference reading operation, and feeds the difference to the decoder to predict the $t$-th word of the response.

\textbf{State Update Operation.} The operation includes SUB and ADD as two sub-operations. Intuitively, when the status of expression surpasses the goal, then the state controller should execute the SUB operation (stands for ``subtract'') to trim the status representation; while when the status of expression is inadequate, then the state controller should use the ADD operation to enhance the status representation. Technically, rather than comparing $\mathcal{M}_i.v_{t-1}$ with $\mathcal{M}_i.g$ and adopting operations accordingly, we propose a soft way to update the state memory panel with SUB and ADD, since (1) it is difficult to identify over-expression or sub-expression by comparing two distributed representations; and (2) the hard way will break differentiablility of the model. Specifically, we define $g_t \in \mathbb{R}^{d\times l}$ as a gate to control the use of SUB or ADD where $g_t(i)\in \mathbb{R}^d$ is the $i$-th element of $g_t$. Let $\Delta_t^{SUB}(i)\in \mathbb{R}^d$  and  $\Delta_t^{ADD}(i)\in \mathbb{R}^d$ be the changes from the SUB operation and the ADD operation respectively, then $\mathcal{M}_i.v_{t-1}$ is updated as
\begin{equation}
\small
\begin{aligned}
& \hat{V}_t(i)  = \mathcal{M}_i.v_{t-1} - g_t(i)\circ\Delta_t^{SUB}(i),\\
& \mathcal{M}_i.v_t  = \hat{V}_t(i) + (1-g_t(i))\circ\Delta_t^{ADD}(i),
\end{aligned}
\end{equation}
where $\circ$ means element-wise multiplication, and $g_t(i)$, $\Delta_t^{SUB}(i)$, and $\Delta_t^{ADD}(i)$ can be defined as
\begin{equation}
\label{g}
\small
g_t(i) = \sigma(W_g\mathcal{S}_t(i) +b_g)
\end{equation}
and
\begin{equation}
\label{addsub}
\small
\begin{bmatrix}
\Delta_t^{SUB}(i)  \\
\Delta_t^{ADD}(i)
\end{bmatrix} = \sigma \left( 
\begin{bmatrix}
W^{SUB} \\
W^{ADD}
\end{bmatrix}\mathcal{S}_t(i) +  \begin{bmatrix}
b^{SUB} \\
b^{ADD}
\end{bmatrix}
\right)
\end{equation}
respectively with $W_g \in \mathbb{R}^{d\times d}$, $b_g \in \mathbb{R}^d$, $W^{\{SUB,ADD\}}\in \mathbb{R}^{d\times 3d}$, and $b^{\{SUB,ADD\}}\in \mathbb{R}^d$ parameters. $\mathcal{S}_t(i)=\mathcal{M}_i.k \oplus \mathcal{M}_i.v_{t-1} \oplus s_t$ where $\oplus$ is a concatenation operator.

\textbf{Difference Reading Operation.} For each variable in the meta-word $M$,  the operation represents the difference between the status of expression and the goal of expression as a vector, and then applies an attention mechanism to the vectors to indicate the decoder the importance of variables in generation of the next word.  Formally, suppose that $d_i^t \in \mathbb{R}^{2d}$ is the difference vector for $m_i \in M$ at step $t$, then $d_i^t$ is defined as
\begin{equation}
%\vspace{-1mm}
\small
\label{diffvec}
     d_i^t= (\mathcal{M}_i.g - \mathcal{M}_i.v_t) \oplus (\mathcal{M}_i.g \circ \mathcal{M}_i.v_t).
\end{equation}
With $(d_1^t,\ldots,d_l^t)$ as a difference memory, the difference reading operation then takes $s_t$ as a query vector and calculates attention over the memory as  
\begin{equation}
\small
\label{attDiff}
\begin{aligned}
   & o_t = \sum\nolimits_{i=1}^l a_i \cdot (U d_i^t),\\
   & a_i^t = \text{softmax}((s_t)^\top(U d_i^t)),
\end{aligned}
\end{equation}
where $(a_1^t, \ldots, a_l^t)$ are attention weights, and $U \in \mathbb{R}^{d\times d}$ is a parameter.

\subsection{Response Decoding}
In decoding, the hidden state $s_t$ is calculated by $\text{GRU}(s_{t-1}, [e(y_{t-1}) \oplus C_t])$, where $e(y_{t-1}) \in \mathbb{R}^d$ is the embedding of the word predicted at step $t-1$, and $C_t$ is a context vector obtained from attention over the hidden states of the input message $X$ given by the biGRU based encoder. Let $H_X = (h_{X,1},\dots,h_{X,T_{x}})$ be the hidden states of $X$, then $C_t$ is calculated via 
\begin{equation}
\small
\label{contextvec}
\begin{aligned}
& C_t  = \sum\nolimits_{j=1}^{T_{x}} \alpha_{t,j}h_{X,j} \\
& \alpha_{t,j} = \frac{exp(e_{t,j})}{\sum_{k=1}^{T_{x}}exp(e_{t,k})},\\
& e_{t,j} = U_d^\top\tanh{(W_{s}s_{t-1} + W_{h}h_{X,j} + b_d)},
\end{aligned}
\end{equation}
where $U_d$, $W_{s}$, $W_{h}$, and $b_d$ are parameters, and $s_{t-1}$ is the hidden state of the decoder at step $t-1$.

With the hidden state $s_t$ and the distance vector $o_t$ returned by the state controller, the probability distribution for predicting the $t$-th word of the response is given by 
\begin{equation}
\small
\label{yprob}
    p(y_{t}) = \text{softmax}(W_p[e(y_{t}) \oplus o_t \oplus s_t ]+b_p),
\end{equation}
where $y_t$ is the $t$-th word of the response with $e(y_{t})$ its embedding, and $W_p$ and $b_p$ are parameters.

\section{Learning Method}
To perform online response generation with meta-words, we need to (1) estimate parameters of GTMES2S by minimizing a loss function; and (2) learn a model to predict meta-words for online messages. 
\subsection{Loss for Model Learning} 
The first loss item is the negative log likelihood (NLL) of $\mathcal{D}$, which is formulated as
%\vspace{-1mm}
\begin{equation}
%\vspace{-1mm}
\small
\label{obj1}
    \mathcal{L}_{NLL} (\Theta) =  -\frac{1}{N}\sum\nolimits_{i=1}^N \log  P(Y_i | X_i, M_i),
\end{equation}
where $\Theta$ is the set of parameters of GTMES2S. By minimizing NLL, the supervision signals in $\mathcal{D}$ may not sufficiently flow to GTMN, as GTMN is nested within response decoding. Thus, besides NLL, we propose a state update loss that directly supervises the learning of GTMN with  $\mathcal{D}$. The idea is to minimize the distance between the ground truth status of meta-word expression and the status stored in the state memory panel. Suppose that $y_{1:t}$ is the segment of response $Y$ generated until step $t$, then $\forall m_i \in M$, we consider two cases: (1) $\exists \mathcal{F}_i(\cdot)$ that $\mathcal{F}_i(y_{1:t})$ maps $y_{1:t}$ to the space of $m_i.v$. As an example, response length belongs to this case with $\mathcal{F}_i(y_{1:t})=t$; (2) it is hard to define an $\mathcal{F}_i(\cdot)$ that can map $y_{1:t}$ to the space of $m_i.v$. For instance, dialogue acts belong to this case since it is often difficult to judge the dialogue act from part of a response. For case (1), we define the state update loss as
\begin{equation}
\small
\label{sloss1}
\mathcal{L}_{SU}^1(m_i)=\sum\nolimits_{t=1}^T\lVert \mathcal{M}_{i}.v_t - Rep(\mathcal{F}_i(y_{1:t}))\rVert,
\end{equation}
where $T$ is the length of $Y$ and $\lVert \cdot \rVert$ refers to $L_2$ norm. For case (2), the loss is defined as
\begin{equation}
\small
\label{sloss2}
\mathcal{L}_{SU}^2(m_i)=\lVert \mathcal{M}_{i}.v_T - Rep(m_{i}.v)\rVert.
\end{equation}
The full state update loss $\mathcal{L}_{SU}(\Theta)$ for $\mathcal{D}$ is then given by
\begin{equation}
\small
\label{sloss}
\sum_{i=1}^N \sum_{j=1}^l \mathbb{I}[m_{i,j} \in \mathcal{C}_1] \mathcal{L}_{SU}^1(m_{i,j}) + \mathbb{I}[m_{i,j} \in \mathcal{C}_2] \mathcal{L}_{SU}^2(m_{i,j}),
\end{equation}
where $\mathcal{C}_1$ and $\mathcal{C}_2$ represent sets of variables belonging to case (1) and case (2) respectively, and $\mathbb{I}(\cdot)$ is an indicator function. The loss function for learning of GTMES2S is finally defined by
\begin{equation}
\small
\label{fullobj}
    \mathcal{L}(\Theta) =   \mathcal{L}_{NLL}(\Theta)  +\lambda \mathcal{L}_{SU}(\Theta), 
\end{equation}
where $\lambda$ acts as a trade-off between the two items. 

\subsection{Meta-word Prediction} 
We assume that values of meta-words are given beforehand. In training, the values can be extracted from ground truth. In test, however, since only a message is available, we propose sampling values of a meta-word for the message from probability distributions estimated from $\{(X_i,M_i)\}_{i=1}^N \subset \mathcal{D}$. The sampling approach not only provides meta-words to GTMNES2S, but also keeps meta-words diverse for similar messages. Formally, let $h_X^p$ be the last hidden state of a message $X$ processed by a biGRU, then $\forall m_i \in M$, we assume that $m_i.v$ obeys a multinomial distribution with the probability $\vec{p}_i$ parameterized as $\text{softmax}(W^{mul}_i h_X^p+b^{mul}_i)$, if $m_i.t=c$; otherwise, $m_i.v$ obeys a normal distribution with $\mu_i$ and $\log(\sigma^2_i)$ parameterized as $W^{\mu}_i h_X^p +b^{\mu}_i$ and $W^{\sigma}_i h_X^p +b^{\sigma}_i$ respectively.  In distribution estimation, we assume that variables in a meta-word are independent, and jointly maximize the log likelihood of $\{(M_i|X_i)\}_{i=1}^N$ and the entropy of the distributions as regularization.

%\vspace{-5mm}
\section{Experiments}
We test GTMNES2S on two large-scale datasets.

%\vspace{-1mm}
\subsection{Datasets}
We mine $10$ million message-response pairs from Twitter FireHose, covering 2-month period from June 2016 to July 2016, and sample $10$ million pairs from the full Reddit data\footnote{\url{https://redd.it/3bxlg7}}. As pre-processing, we remove duplicate pairs, pairs with a message or a response having more than $30$ words, and messages that correspond to more than $20$ responses to prevent them from dominating learning.  After that, there are $4,759,823$ pairs left for Twitter and $4,246,789$ pairs left for Reddit.  On average, each message contains $10.78$ words in the Twitter data and  $12.96$ words in the Reddit data. The average lengths of responses in the Twitter data and the Reddit data are $11.03$ and $12.75$ respectively. From the pairs after pre-processing, we randomly sample $10$k pairs as a validation set and $10$k pairs as a test set for each data, and make sure that there is no overlap between the two sets. After excluding pairs in the validation sets and the test sets, the left pairs are used for model training. The test sets are built for calculating automatic metrics. Besides, we randomly sample $1000$ distinct messages from each of the two test sets and recruit human annotators to judge the quality of responses generated for these messages. For both the Twitter data and the Reddit data, top $30,000$ most frequent words in messages and responses in the training sets are kept as message vocabularies and response vocabularies. In the Twitter data, the message vocabulary and the response vocabulary cover $99.17$\% and $98.67$\% words appearing in messages and responses respectively. The two ratios are $99.52$\% and $98.8$\% respectively in the Reddit data. Other words are marked as ``UNK''. 

%\vspace{-1mm}
\subsection{Meta-word Construction}\label{metacons}
As a showcase of the framework of GTMNES2S, we consider the following variables as a meta-word: (1) \textbf{Response Length (RL)}: number of words and punctuation marks in a response. We restrict the range of the variable in $\{1,\ldots, 25\}$ (i.e., responses longer than $25$ are normalized as $25$), and treat it as a categorical variable. (2) \textbf{Dialog Act (DA)}: we employ the $42$ dialogue acts based on the DAMSL annotation scheme \cite{core1997coding}. The dialogue act of a given response is obtained by the state-of-the-art dialogue act classifier in \cite{liu2017using} learned from the Switchboard (SW) 1 Release 2 Corpus \cite{godfrey1997switchboard}. DA is a categorical variable. (3) \textbf{Multiple Utterances (MU)}: if a response is made up of multiple utterances. We split a  response as utterances according to ``.'', ``?'' and ``!'', and remove utterances that are less than $3$ words. The variable is ``true'' if there are more than $1$ utterance left, otherwise it is ``false''. (4) \textbf{Copy Ratio (CR)}: inspired by COPY-NET \cite{gu2016incorporating} which indicates that humans may repeat entity names or even long phrases in conversation, we incorporate a ``copy mechanism'' into our model by using copy ratio as a soft implementation of COPY-NET. We compute the ratio of unigrams shared by a message and its response (divided by the length of the response) with stop words and top $1000$ most frequent words in training excluded. CR is a real-valued variable. (5) \textbf{Specificity (S)}: following SC-Seq2Seq \cite{zhang2018generating}, we calculate normalized inverse word frequency as a specificity variable. The variable is real-valued.  Among the five variables, RL, CR, and S correspond to the state update loss given by Equation (\ref{sloss1}), and others correspond to Equation (\ref{sloss2}). 

\subsection{Baselines}
We compare GTMNES2S with the following baseline models: (1) \textbf{MMI-bidi}: the sequence-to-sequence model with response re-ranking in \cite{li2015diversity} learned by a maximum mutual information objective; (2) \textbf{SC-Seq2Seq}: the specificity controlled Seq2Seq model in \cite{zhang2018generating}; (3) \textbf{kg-CVAE}: the knowledge-guided conditional variational autoencoders in \cite{zhao2017learning}; and (4) \textbf{CT}: the conditional training method in \cite{see2019makes} that feeds the embedding of pre-defined response attributes to the decoder of a sequence-to-sequence model. Among the baselines, CT exploits the same attributes as GTMNES2S, SC-Seq2Seq utilizes specificity, and kg-CVAE leverages dialogue acts. All models are implemented with the recommended parameter configurations in the existing papers, where for kg-CVAE, we use the code shared at \url{https://github.com/snakeztc/NeuralDialog-CVAE}, and for other models without officially published code, we code with TensorFlow.  Besides the baselines, we also compare GTMNES2E learned from the full loss given by Equation (\ref{fullobj}) with a variant learned only from the NLL loss, in order to check the effect of the proposed state update loss. We denote the variant as GTMNES2S \textit{w/o} SU.

\subsection{Evaluation Metrics}
We conduct both automatic evaluation and human evaluation. In terms of automatic ways, we evaluate models from four aspects: relevance, diversity, accuracy of one-to-many modeling, and accuracy of meta-word expression. For relevance, besides BLEU \cite{papineni2002bleu}, we follow \cite{serban2017hierarchical} and employ Embedding Average (Average), Embedding Extrema (Extrema), Embedding Greedy (Greedy) as metrics. To evaluate diversity, we follow \cite{li2015diversity} and use Distinct-1 (Dist1) and Distinct-2 (Dist2) as metrics which are calculated as the ratios of distinct unigrams and bigrams in the generated responses. For accuracy of one-to-many modeling, we utilize A-bow precision (A-prec), A-bow recall (A-rec), E-bow precision (E-prec), and E-bow recall (E-rec) proposed in \cite{zhao2017learning} as metrics. For accuracy of meta-word expression, we measure accuracy for categorical variables and square deviation for real-valued variables. Metrics of relevance, diversity, and accuracy of meta-word expression are calculated on the $10$k test data based on top $1$ responses from beam search. To measure the accuracy of meta-word expression for a generated response, we extract values of the meta-word of the response with the methods described in Section \ref{metacons}, and compare these values with the oracle ones sampled from distributions.  Metrics of accuracy of one-to-many modeling require a test message to have multiple reference responses. Thus, we filter the test sets by picking out messages that have at least $2$ responses, and form two subsets with $166$ messages for Twitter and $135$ messages for Reddit respectively. On average, each message corresponds to $2.8$ responses in the Twitter data and $2.92$ responses in the Reddit data. For each message, $10$ responses from a model are used for evaluation. In kg-CVAE, we follow \cite{zhao2017learning} and sample $10$ times from the latent variable; in SC-Seq2Seq, we vary the specificity in $\{0.1,0.2,\ldots,1\}$; and in both CT and GTMNES2S, we sample $10$ times from the distributions. Top $1$ response from beam search under each sampling or specificity setting are collected as the set for evaluation.   

In terms of human evaluation, we recruit $3$ native speakers to label top $1$ responses of beam search from different models. Responses from all models for all the $1000$ test messages in both data are pooled, randomly shuffled, and presented to each of the annotators.  The annotators judge the quality of the responses according to the following criteria: \textbf{+2}: the response is not only relevant and natural, but also informative and interesting; \textbf{+1}: the response can be used as a reply, but might not be informative enough (e.g.,“Yes, I see” etc.); \textbf{0}: the response makes no sense, is irrelevant, or is grammatically broken. Each response receives $3$ labels. Agreements among the annotators are measured by Fleiss' kappa \cite{fleiss1973equivalence}.

\begin{table*}[t!]  \small
%\vspace{-2.5mm}
	\centering
	\resizebox{0.88\linewidth}{!}{
    \begin{tabular}{c|r|c|c|c|c|c|c|c|c|c|c}
    \hline
       \multirow{2}{*}{Dataset} & \multirow{2}{*}{Models} &   \multicolumn{4}{c|}{\textbf{Relevance}} 	&   \multicolumn{2}{c|}{\textbf{Diversity}} & \multicolumn{4}{c}{\textbf{One-to-Many}} \\\cline{3-12}
                      &    &  BLEU  & Average & Greedy & Extreme & Dist1 & Dist2  & A-prec & A-rec & E-prec & E-rec\\ 
    \hline
	\multirow{6}{*}{Twitter} &  MMI-bidi  & 2.92 & 0.787 & 0.181 & 0.394 & 6.35 & 20.6 & 0.853 & 0.810 & 0.601 & 0.554\\
	                         &  kg-CVAE     & 1.83 & 0.766 & 0.175 & 0.373 & 8.65 & 29.7 & 0.862 & 0.822 & 0.597 & 0.545\\
                             &  SC-Seq2Seq    & 2.57 & 0.776 & 0.182 & 0.387 & 6.87 & 22.5 & 0.857 & 0.815 & 0.594 & 0.551\\
                             &  CT     & 3.32 & 0.792 & 0.181 & 0.402 & 8.04 & 26.9 & 0.859 & 0.813 & 0.596 & 0.550\\
                             &  GTMNES2S \textit{w/o} SU     & 3.25 & 0.793 & 0.183 & 0.405 &  7.59 & 28.4 & 0.861 & 0.819 & 0.598 & 0.554\\
                             
                             &  GTMNES2S     & 3.39 & \bf0.810 & 0.182 & \bf0.413 &  8.41 & \bf30.5 & \bf0.886 & \bf0.839 & \bf0.610 & 0.560\\
                           
    \hline
    \multirow{6}{*}{Reddit}  &  MMI-bidi  &  1.82  & 0.752 & 0.171 & 0.369 & 6.12 & 20.3 & 0.821 & 0.775 & 0.587 & 0.542\\
                             &  kg-CVAE     &  1.89  & 0.745 & 0.171 & 0.357 & 8.47 & 28.7 & 0.827 & 0.781 &0.583 & 0.531\\
                             &  SC-Seq2Seq    &  1.95  & 0.752 & 0.176 & 0.362 & 5.94 & 19.2 & 0.823 & 0.778 & 0.581 & 0.536\\
                             &  CT      &  2.43  & 0.751 & 0.172 & 0.383 &  8.62 & 33.4 &  0.827 & 0.783 & 0.587 & 0.540\\
                             &  GTMNES2S \textit{w/o} SU      &  2.75  & 0.757 & 0.174 & 0.382 &  8.47 & 32.6 & 0.832 & 0.791 & 0.594 & 0.548\\
      
                             &  GTMNES2S    &  \bf2.95  & \bf0.760 & 0.172 & 0.386 &  \bf10.35 & \bf36.3 & \bf0.841 & \bf0.795 & \bf0.602 & \bf0.554\\

    \bottomrule
	\end{tabular}
	}
	%\vspace{-2.5mm}
	\caption{Results on relevance, diversity, and accuracy of one-to-many modeling. Numbers in bold mean that improvement over the best baseline is statistically significant (t-test, p-value $< 0.01$).}
    \label{tab:auto-metrics}
    %\vspace{-2mm}
\end{table*}

%\vspace{-1mm}
\subsection{Implementation Details}
In test, we fix the specificity variable as $0.5$ in SC-Seq2Seq, since in \cite{zhang2018learning}, the authors conclude that the model achieves the best overall performance under the setting. For kg-CVAE, we follow \cite{zhao2017learning} and predict a dialogue act for a message with an MLP.  GTMNES2S and CT leverage the same set of attributes. Thus, for fair comparison, we let them exploit the same sampled values in generation.  In GTMNES2S, the size of hidden units of the encoder and the decoder, and the size of the vectors in memory cells (i.e., $d$) are $512$. Word embedding is randomly initialized with a size of $512$. We adopt the Adadelta algorithm \cite{zeiler2012adadelta} in optimization with a batch size $200$. Gradients are clipped when their norms exceed $5$. We stop training when the perplexity of a model on the validation data does not drop in two consecutive epochs.  Beam sizes are $200$ in MMI-bidi (i.e., the size used in \cite{li2015diversity}) and $5$ in other models.

\subsection{Evaluation Results}

\begin{table*}[t!] 
\centering
\resizebox{0.75\linewidth}{!}{
    \centering	
    \begin{tabular}{c r c c c c c c
    }
        \toprule
        Dataset &   Metaword & Type & SC-Seq2Seq&kg-CVAE
        
        & CT & GTMNES2S \textit{w/o} SU & GTMNES2S\\ 
        \midrule
        \multirow{5}{*}{Twitter}
                            & RL & c & -&-
                  
                  & 97\% & 95.6\%&\bf98.6\% \\
                                & DA &c & -& 58.2\%      
                  & 60.9\% &61.2\% &\bf62.6\% \\
                                & MU &c &-&- 
                  & 98.8\% & 99.5\%&99.4\%\\
                                & CR &r &-&-   
                  & 0.176 & 0.178&\bf0.164 \\
                                & S & r &0.195&- 
                  
                  & 0.130 & 0.158 &\bf0.103 \\
        \midrule
        \multirow{5}{*}{Reddit}   & RL & c &-&- 
        & 94.5\% & 95.1\% &\bf96.7\%\\ 
                                & DA &c &-&55.7\%
                  
                  & 59.9\% & 55.9\%&\bf61.2\% \\
                                & MU &c&-&-  
                  & 99.2\% & 98.7\%&99.4\%\\
                                & CR &r&-&- 
                  & 0.247 & 0.253&\bf0.236 \\
                                & S & r &0.143&- 
                  & 0.118 & 0.112 &\bf0.084 \\
    \bottomrule
    \end{tabular}
    }
    %\vspace{-2mm}
    \caption{Results on accuracy of meta-word expression. Numbers in bold mean that improvement over the best baseline is statistically significant (t-test, p-value $< 0.01$).}		
    %\vspace{-6mm}
    \label{tab:accexpression}
\end{table*}

Table \ref{tab:auto-metrics} and Table \ref{tab:accexpression} report evaluation results on automatic metrics. On most of the metrics, GTMNES2S outperforms all baseline methods, and the improvements are significant in a statistical sense (t-test, p-value $< 0.01$). The results demonstrate that with meta-words, our model can represent the relationship between messages and responses in a more effective and more accurate way, and thus can generate more diverse responses without sacrifice on relevance. Despite leveraging the same attributes for response generation, GTMNES2S achieves better accuracy than CT on both one-to-many modeling and meta-word expression, indicating the advantages of the dynamic control strategy over the static control strategy, as we have analyzed at the beginning of Section \ref{GTMN}. Without the state update loss, there is significant performance drop for GTMNES2S. The results verified the effect of the proposed loss in learning. Table \ref{tab:human-evaluation} summarizes human evaluation results. Compared with the baseline methods and the variant, the full GTMNES2S model can generate much more excellent responses (labeled as ``2'') and much fewer inferior responses (labeled as ``0''). Kappa values of all models exceed $0.6$, indicating substantial agreement over all annotators. The results further demonstrate the value of the proposed model for real human-machine conversation. kg-CVAE gives more informative responses, and also more bad responses than MMI-bidi and SC-Seq2Seq. Together with the contradiction on diversity and relevance in Table \ref{tab:auto-metrics}, the results indicate that latent variable is a double-bladed sword: the randomness may bring interesting content to responses and may also make responses out of control. On the other hand, there are no random variables in our model, and thus, it can enjoy a well-trained language model.

\begin{table}[t!]
\centering
\resizebox{\linewidth}{!}{
    %\vspace{-10mm}
    \centering	
    \begin{tabular}{c r c c c c c}
        \toprule
        Dataset &   Models & 2 & 1 & 0 & Avg&kappa\\ 
        \midrule
        \multirow{6}{*}{Twitter}
                            & MMI-bidi&16.6\% & 51.7\% & 31.7\% & 0.85&0.65\\
                            & kg-CVAE & 23.1\% & 40.9\% & 36\% & 0.87&0.78 \\ 
                            & SC-Seq2Seq&21.2\% & 48.5\% & 30.3\% & 0.91& 0.61\\
                                  & CT & 27.6\%  & 38.4\% & 34\% & 0.94 &0.71 \\
                                   & GTMNES2S \textit{w/o} SU & 27\%  & 39.1\% & 33.9\% & 0.93&0.64 \\
                                & GTMNES2S & 33.2\%  & 37.7\% & 29.1\% & 1.04&0.71 \\
                                
        \midrule
        \multirow{6}{*}{Reddit}   & MMI-bidi & 4.4\% & 58.1\% & 37.5\% & 0.67&0.79\\
                                  & kg-CVAE & 13.7\% & 44.6\% & 41.7\% & 0.72&0.68 \\ 
                                  & SC-Seq2Seq & 9.9\% & 51.2\% & 38.9\% & 0.71& 0.78\\
                          & CT & 16.5\%  & 48.2\% & 35.3\% & 0.81&0.73 \\
                                   & GTMNES2S \textit{w/o} SU & 15.7\%  & 47.3\% & 37\% & 0.79&0.66 \\
                                & GTMNES2S & 19.2\%  & 47.5\% & 33.3\% & 0.86&0.76 \\

    \bottomrule
    \end{tabular}
    }
    %\vspace{-2mm}
    \caption{Results on the human evaluation. Ratios are calculated by combining labels from the three judges.}		
    \label{tab:human-evaluation}
%\vspace{-7mm}
\end{table}

%\vspace{-1.5mm}
\subsection{Discussions}
In this section, we examine effect of different attributes by adding them one by one to the generation model. Besides, we also illustrate how GTMNES2S tracks attribute expression in response generation with test examples. 

\begin{table}[t!]
\centering
\resizebox{\linewidth}{!}{
    \begin{tabular}{ccccccc}
        \toprule
        \multirow{2}{*}{Dataset} &   Multiple  & Dialog  & \multirow{2}{*}{Length} & Copy  & \multirow{2}{*}{Specificity} & \multirow{2}{*}{PPL} \\
        & utterances & Act & & Ratio & &  \\ \hline
        \multirow{6}{*}{Twitter} & $\times$ & $\times$ & $\times$ & $\times$ & $\times$ &70.19 \\
      & $\checkmark$ & $\times$ & $\times$ & $\times$ & $\times$ &67.23 \\
      & $\checkmark$ & $\checkmark$ & $\times$ & $\times$ & $\times$ &62.13 \\
      & $\checkmark$ & $\checkmark$ & $\checkmark$ & $\times$ & $\times$ &50.36 \\
      & $\checkmark$ & $\checkmark$ & $\checkmark$ & $\checkmark$ & $\times$ &42.05 \\
      & $\checkmark$ & $\checkmark$ & $\checkmark$ & $\checkmark$ & $\checkmark$ &38.57 \\
        \midrule
    
    \multirow{6}{*}{Reddit}  & $\times$ & $\times$ & $\times$ & $\times$ & $\times$ & 72.43 \\
      & $\checkmark$ & $\times$ & $\times$ & $\times$ & $\times$ &65.17 \\
      & $\checkmark$ & $\checkmark$ & $\times$ & $\times$ & $\times$ & 61.92 \\
      & $\checkmark$ & $\checkmark$ & $\checkmark$ & $\times$ & $\times$ & 49.67 \\
      & $\checkmark$ & $\checkmark$ & $\checkmark$ & $\checkmark$ & $\times$ &41.78 \\
      & $\checkmark$ & $\checkmark$ & $\checkmark$ & $\checkmark$ & $\checkmark$ &37.96 \\
    \bottomrule
    \end{tabular}
}
\caption{Contribution of different attributes.}		
\label{tab:combination}
\end{table}

%\vspace{-2mm}
\paragraph{Contribution of attributes.} Table \ref{tab:combination} shows perplexity (PPL) of GTMNES2S with different sets of attributes on the validation data. We can see that the more attributes are involved in learning, the lower PPL we can get. By leveraging all the $5$ attributes, we can reduce almost $50$\% PPL from the vanilla encoder-decoder model (i.e., the one without any attributes). The results not only indicate the contribution of different attributes to model fitting, but also inspire us the potential of the proposed framework, since it allows further improvement with more well designed attributes involved. 

\begin{figure}[t!]
    \centering
    %\vspace{-4mm}
    \includegraphics[width=0.485\textwidth]{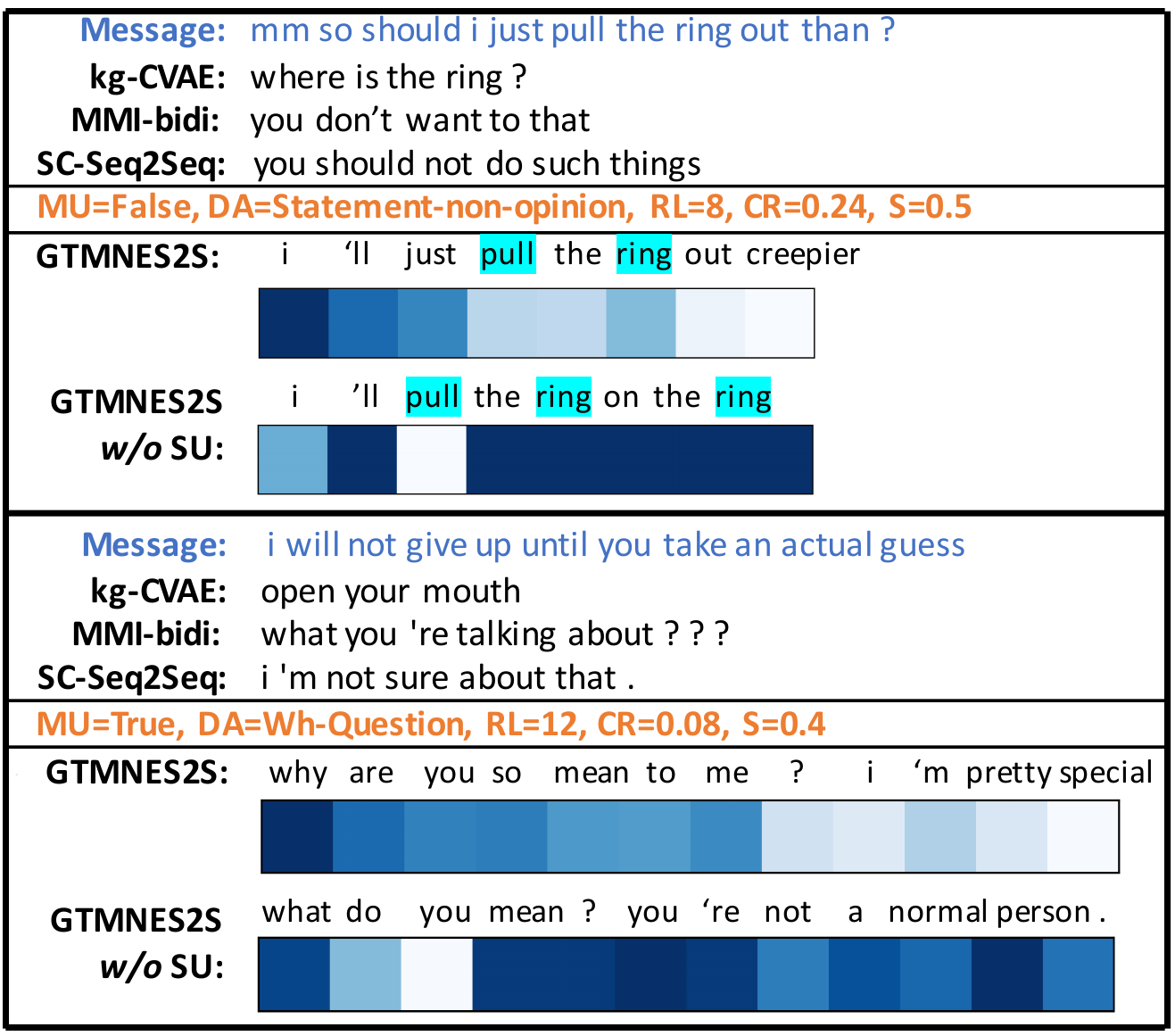}
    %\vspace{-3mm}
    \caption{Examples of response generation from the Twitter test data. \textbf{Up}: the heat map is defined by $\lVert \mathcal{M}_i.v_t - \mathcal{M}_{i}.g\rVert$ normalized to $[0,1]$, where $\mathcal{M}_i$ refers to CR. \textbf{Below}: $\mathcal{M}_i$ in the heat map refers to MU.}
    %\vspace{-7mm}
    \label{fig:case}
\end{figure}

%\vspace{-1.5mm}
\paragraph{Case Study.} Figure \ref{fig:case} illustrates how our model controls attributes of responses with the goal tracking mechanism, where distance between the value of a memory cell (i.e., $\mathcal{M}_i.v_t$) during generation and the goal of the memory cell (i.e., $\mathcal{M}_i.g$) is visualized via heat maps. In the first example, the full model gradually reduces the distance between the value and the goal of copy ratio expression with the generation process moving on. As a result, it  just copies ``pull the ring out'' from the message, which makes the response informative and coherent. On the other hand, without the state update loss, GTMNES2S \textit{w/o} SU makes a mistake by copying ``ring'' twice, and the distance between the value and the goal is out of control. In the second example, we visualize the expression of MU, a categorical attribute. Compared with real-valued attributes, categorical attributes are easier to express. Therefore, both the full model and GTMNES2S \textit{w/o} SU successfully generate a response with multiple utterances, although the distance between the value and the goal of MU expression in GTMNES2S \textit{w/o} SU is still in a mess.

%\vspace{-2mm}
\section{Conclusions}
%\vspace{-2mm}
We present a goal-tracking memory enhanced sequence-to-sequence model for open domain response generation with meta-words which explicitly define characteristics of responses. Evaluation results on two datasets indicate that our model significantly outperforms several state-of-the-art generative architectures in terms of both response quality and accuracy of meta-word expression.  

\bibliography{acl2019}

\begin{thebibliography}{32}
\expandafter\ifx\csname natexlab\endcsname\relax\def\natexlab#1{#1}\fi

\bibitem[{Chung et~al.(2014)Chung, Gulcehre, Cho, and
  Bengio}]{chung2014empirical}
Junyoung Chung, Caglar Gulcehre, KyungHyun Cho, and Yoshua Bengio. 2014.
\newblock Empirical evaluation of gated recurrent neural networks on sequence
  modeling.
\newblock \emph{arXiv preprint arXiv:1412.3555}.

\bibitem[{Core and Allen(1997)}]{core1997coding}
Mark~G Core and James Allen. 1997.
\newblock Coding dialogs with the damsl annotation scheme.
\newblock In \emph{AAAI fall symposium on communicative action in humans and
  machines}, volume~56. Boston, MA.

\bibitem[{Fleiss and Cohen(1973)}]{fleiss1973equivalence}
Joseph~L Fleiss and Jacob Cohen. 1973.
\newblock The equivalence of weighted kappa and the intraclass correlation
  coefficient as measures of reliability.
\newblock \emph{Educational and psychological measurement}, 33(3):613--619.

\bibitem[{Godfrey and Holliman(1997)}]{godfrey1997switchboard}
John~J Godfrey and Edward Holliman. 1997.
\newblock Switchboard-1 release 2.
\newblock \emph{Linguistic Data Consortium, Philadelphia}, 926:927.

\bibitem[{Gu et~al.(2016)Gu, Lu, Li, and Li}]{gu2016incorporating}
Jiatao Gu, Zhengdong Lu, Hang Li, and Victor~OK Li. 2016.
\newblock Incorporating copying mechanism in sequence-to-sequence learning.
\newblock \emph{arXiv preprint arXiv:1603.06393}.

\bibitem[{Hu et~al.(2017)Hu, Yang, Liang, Salakhutdinov, and
  Xing}]{hu2017toward}
Zhiting Hu, Zichao Yang, Xiaodan Liang, Ruslan Salakhutdinov, and Eric~P Xing.
  2017.
\newblock Toward controlled generation of text.
\newblock In \emph{International Conference on Machine Learning}, pages
  1587--1596.

\bibitem[{Li et~al.(2015)Li, Galley, Brockett, Gao, and
  Dolan}]{li2015diversity}
Jiwei Li, Michel Galley, Chris Brockett, Jianfeng Gao, and Bill Dolan. 2015.
\newblock A diversity-promoting objective function for neural conversation
  models.
\newblock In \emph{Proceedings of the 2016 Conference of the North American
  Chapter of the Association for Computational Linguistics: Human Language
  Technologies}, pages 110--119.

\bibitem[{Li et~al.(2016)Li, Galley, Brockett, Spithourakis, Gao, and
  Dolan}]{li2016persona}
Jiwei Li, Michel Galley, Chris Brockett, Georgios Spithourakis, Jianfeng Gao,
  and Bill Dolan. 2016.
\newblock A persona-based neural conversation model.
\newblock In \emph{Proceedings of the 54th Annual Meeting of the Association
  for Computational Linguistics}, pages 994--1003.

\bibitem[{Liu et~al.(2017)Liu, Han, Tan, and Lei}]{liu2017using}
Yang Liu, Kun Han, Zhao Tan, and Yun Lei. 2017.
\newblock Using context information for dialog act classification in dnn
  framework.
\newblock In \emph{Proceedings of the 2017 Conference on Empirical Methods in
  Natural Language Processing}, pages 2170--2178.

\bibitem[{Logeswaran et~al.(2018)Logeswaran, Lee, and
  Bengio}]{logeswaran2018content}
Lajanugen Logeswaran, Honglak Lee, and Samy Bengio. 2018.
\newblock Content preserving text generation with attribute controls.
\newblock In \emph{Advances in Neural Information Processing Systems}, pages
  5108--5118.

\bibitem[{Mou et~al.(2016)Mou, Song, Yan, Li, Zhang, and Jin}]{mou2016sequence}
Lili Mou, Yiping Song, Rui Yan, Ge~Li, Lu~Zhang, and Zhi Jin. 2016.
\newblock Sequence to backward and forward sequences: A content-introducing
  approach to generative short-text conversation.
\newblock In \emph{Proceedings of the 26th International Conference on
  Computational Linguistics: Technical Papers}, pages 3349--3358.

\bibitem[{Papineni et~al.(2002)Papineni, Roukos, Ward, and
  Zhu}]{papineni2002bleu}
Kishore Papineni, Salim Roukos, Todd Ward, and Wei-Jing Zhu. 2002.
\newblock \href {http://aclweb.org/anthology/P02-1040} {Bleu: a method for
  automatic evaluation of machine translation}.
\newblock In \emph{Proceedings of the 40th Annual Meeting of the Association
  for Computational Linguistics}, pages 311--318.

\bibitem[{Park et~al.(2018)Park, Cho, and Kim}]{park2018hierarchical}
Yookoon Park, Jaemin Cho, and Gunhee Kim. 2018.
\newblock A hierarchical latent structure for variational conversation
  modeling.
\newblock In \emph{Proceedings of the 2018 Conference of the North American
  Chapter of the Association for Computational Linguistics: Human Language
  Technologies, Volume 1 (Long Papers)}, volume~1, pages 1792--1801.

\bibitem[{See et~al.(2019)See, Roller, Kiela, and Weston}]{see2019makes}
Abigail See, Stephen Roller, Douwe Kiela, and Jason Weston. 2019.
\newblock What makes a good conversation? how controllable attributes affect
  human judgments.
\newblock \emph{arXiv preprint arXiv:1902.08654}.

\bibitem[{Serban et~al.(2016)Serban, Sordoni, Bengio, Courville, and
  Pineau}]{serban2015building}
Iulian~Vlad Serban, Alessandro Sordoni, Yoshua Bengio, Aaron~C. Courville, and
  Joelle Pineau. 2016.
\newblock End-to-end dialogue systems using generative hierarchical neural
  network models.
\newblock In \emph{AAAI}, pages 3776--3784.

\bibitem[{Serban et~al.(2017)Serban, Sordoni, Lowe, Charlin, Pineau, Courville,
  and Bengio}]{serban2017hierarchical}
Iulian~Vlad Serban, Alessandro Sordoni, Ryan Lowe, Laurent Charlin, Joelle
  Pineau, Aaron~C Courville, and Yoshua Bengio. 2017.
\newblock A hierarchical latent variable encoder-decoder model for generating
  dialogues.
\newblock In \emph{AAAI}, pages 3295--3301.

\bibitem[{Shang et~al.(2015)Shang, Lu, and Li}]{shangL2015neural}
Lifeng Shang, Zhengdong Lu, and Hang Li. 2015.
\newblock Neural responding machine for short-text conversation.
\newblock In \emph{ACL}, pages 1577--1586.

\bibitem[{Shum et~al.(2018)Shum, He, and Li}]{shum2018eliza}
Heung{-}Yeung Shum, Xiaodong He, and Di~Li. 2018.
\newblock \href {https://doi.org/10.1631/FITEE.1700826} {From eliza to xiaoice:
  Challenges and opportunities with social chatbots}.
\newblock \emph{Frontiers of Information Technology {\&} Electronic
  Engineering}, 19(1):10--26.

\bibitem[{Sutskever et~al.(2014)Sutskever, Vinyals, and
  Le}]{sutskever2014sequence}
Ilya Sutskever, Oriol Vinyals, and Quoc~V. Le. 2014.
\newblock \href {http://dl.acm.org/citation.cfm?id=2969033.2969173} {Sequence
  to sequence learning with neural networks}.
\newblock In \emph{Proceedings of the 27th International Conference on Neural
  Information Processing Systems - Volume 2}, NIPS'14, pages 3104--3112,
  Cambridge, MA, USA. MIT Press.

\bibitem[{Tao et~al.(2018)Tao, Gao, Shang, Wu, Zhao, and Yan}]{tao2018get}
Chongyang Tao, Shen Gao, Mingyue Shang, Wei Wu, Dongyan Zhao, and Rui Yan.
  2018.
\newblock Get the point of my utterance! learning towards effective responses
  with multi-head attention mechanism.
\newblock In \emph{IJCAI}, pages 4418--4424.

\bibitem[{Vinyals and Le(2015)}]{vinyals2015neural}
Oriol Vinyals and Quoc Le. 2015.
\newblock A neural conversational model.
\newblock \emph{arXiv preprint arXiv:1506.05869}.

\bibitem[{Wang et~al.(2017)Wang, Jojic, Brockett, and
  Nyberg}]{wang2017steering}
Di~Wang, Nebojsa Jojic, Chris Brockett, and Eric Nyberg. 2017.
\newblock Steering output style and topic in neural response generation.
\newblock In \emph{Proceedings of the 2017 Conference on Empirical Methods in
  Natural Language Processing}, pages 2140--2150.

\bibitem[{Wu et~al.(2018)Wu, Wu, Yang, Xu, Li, and Zhou}]{wu2017neural}
Yu~Wu, Wei Wu, Dejian Yang, Can Xu, Zhoujun Li, and Ming Zhou. 2018.
\newblock Neural response generation with dynamic vocabularies.
\newblock In \emph{AAAI}, pages 5594--5601.

\bibitem[{Xing et~al.(2017)Xing, Wu, Wu, Liu, Huang, Zhou, and
  Ma}]{xing2017topic}
Chen Xing, Wei Wu, Yu~Wu, Jie Liu, Yalou Huang, Ming Zhou, and Wei-Ying Ma.
  2017.
\newblock Topic aware neural response generation.
\newblock In \emph{AAAI}, pages 3351--3357.

\bibitem[{Xing et~al.(2018)Xing, Wu, Wu, Zhou, Huang, and
  Ma}]{xing2017hierarchical}
Chen Xing, Wei Wu, Yu~Wu, Ming Zhou, Yalou Huang, and Wei-Ying Ma. 2018.
\newblock Hierarchical recurrent attention network for response generation.
\newblock In \emph{AAAI}, pages 5610--5617.

\bibitem[{Young et~al.(2013)Young, Gasic, Thomson, and
  Williams}]{young2013pomdp}
Stephanie Young, Milica Gasic, Blaise Thomson, and John~D Williams. 2013.
\newblock Pomdp-based statistical spoken dialog systems: A review.
\newblock \emph{Proceedings of the IEEE}, 101(5):1160--1179.

\bibitem[{Zeiler(2012)}]{zeiler2012adadelta}
Matthew~D Zeiler. 2012.
\newblock Adadelta: an adaptive learning rate method.
\newblock \emph{arXiv preprint arXiv:1212.5701}.

\bibitem[{Zhang et~al.(2018{\natexlab{a}})Zhang, Guo, Fan, Lan, Xu, and
  Cheng}]{zhang2018learning}
Ruqing Zhang, Jiafeng Guo, Yixing Fan, Yanyan Lan, Jun Xu, and Xueqi Cheng.
  2018{\natexlab{a}}.
\newblock Learning to control the specificity in neural response generation.
\newblock In \emph{Proceedings of the 56th Annual Meeting of the Association
  for Computational Linguistics (Volume 1: Long Papers)}, volume~1, pages
  1108--1117.

\bibitem[{Zhang et~al.(2018{\natexlab{b}})Zhang, Galley, Gao, Gan, Li,
  Brockett, and Dolan}]{zhang2018generating}
Yizhe Zhang, Michel Galley, Jianfeng Gao, Zhe Gan, Xiujun Li, Chris Brockett,
  and Bill Dolan. 2018{\natexlab{b}}.
\newblock Generating informative and diverse conversational responses via
  adversarial information maximization.
\newblock In \emph{Advances in Neural Information Processing Systems}, pages
  1815--1825.

\bibitem[{Zhao et~al.(2018)Zhao, Lee, and Eskenazi}]{zhao2018unsupervised}
Tiancheng Zhao, Kyusong Lee, and Maxine Eskenazi. 2018.
\newblock Unsupervised discrete sentence representation learning for
  interpretable neural dialog generation.
\newblock \emph{arXiv preprint arXiv:1804.08069}.

\bibitem[{Zhao et~al.(2017)Zhao, Zhao, and Eskenazi}]{zhao2017learning}
Tiancheng Zhao, Ran Zhao, and Maxine Eskenazi. 2017.
\newblock Learning discourse-level diversity for neural dialog models using
  conditional variational autoencoders.
\newblock In \emph{Proceedings of the 55th Annual Meeting of the Association
  for Computational Linguistics (Volume 1: Long Papers)}, volume~1, pages
  654--664.

\bibitem[{Zhou et~al.(2018)Zhou, Huang, Zhang, Zhu, and
  Liu}]{zhou2017emotional}
Hao Zhou, Minlie Huang, Tianyang Zhang, Xiaoyan Zhu, and Bing Liu. 2018.
\newblock Emotional chatting machine: Emotional conversation generation with
  internal and external memory.
\newblock In \emph{AAAI}, pages 730--738.

\end{thebibliography}
\bibliographystyle{acl_natbib}

\end{document}